\documentclass[11pt]{article}
\usepackage[utf8]{inputenc}
\usepackage[T1]{fontenc}
\usepackage{mathptmx}
\usepackage{courier}
\usepackage[margin=1in]{geometry}
\usepackage{amsmath}
\usepackage{array}
\usepackage{enumitem}
\usepackage{multirow}
\usepackage{url}
\usepackage[numbers,square,comma,sort&compress]{natbib}
\usepackage[hidelinks]{hyperref}
\usepackage[font=small,labelfont=bf,labelsep=colon]{caption}

\newcolumntype{L}[1]{>{\raggedright\arraybackslash}p{#1}}
\newcolumntype{C}[1]{>{\centering\arraybackslash}p{#1}}
\setlist{itemsep=2pt,topsep=2pt,parsep=0pt}
\newenvironment{keywords}{\par\noindent\textbf{Keywords:}~}{\par}
\newenvironment{abbreviations}{\par\noindent\textbf{Abbreviations:}~}{\par}

\title{Trustworthiness Costs of Domain Adaptation in Small Language Models: A Cross-Architecture Empirical Study}
\author{Ramesh B. Paramkusham\\[2pt]
\small Independent Researcher, Prosper, Texas 75078, USA\\
\small Adjunct Faculty, University of the Cumberlands, Williamsburg, KY 40769, USA\\
\small \texttt{rbpdlf@gmail.com}}
\date{September 2026 (v2)}

\begin{document}
\maketitle

\begin{abstract}
Domain adaptation of small language models is a practical strategy for deploying natural language processing systems in high-stakes settings such as healthcare, legal services, and finance, but its effect on trustworthiness remains poorly understood. This paper presents a cross-domain, cross-architecture empirical study of three small language models (TinyLlama 1B, Gemma-2 2B, Llama 3.2 1B), three domains, two training-data conditions (benign and adversarially perturbed), and four fine-tuning strategies (baseline quantised low-rank adaptation, safety-oriented direct preference optimisation, safety-data replay, and task-arithmetic merging of a safety adapter). Trustworthiness is measured by TruthfulQA MC2 (factual calibration) and by a refusal-based attack success rate on the 200 HarmBench standard behaviours (harm susceptibility), across 216 runs with three random seeds. Four findings emerge. First, the cost of baseline adaptation falls mainly on factual calibration and is model-dependent: Gemma-2 2B lost 0.056--0.069 in TruthfulQA MC2 on healthcare and finance, whereas Llama 3.2 1B gained about 0.02 on legal and finance. Second, the attack success rate of the safety-aligned models rose only modestly (mean $+0.033$), most for Gemma-2 2B on healthcare ($+0.103$). Third, adversarially perturbed training data did not systematically worsen either measure. Fourth, no safety-preserving strategy reduced harm susceptibility in the aligned models: preference optimisation was neutral, safety replay raised the attack success rate by 0.096 on average (up to 0.45 in single cells), and a full-weight task-arithmetic merge was near-neutral ($+0.013$) but cost Gemma-2 2B a further 0.04 in TruthfulQA MC2. A typographic flaw in keyword-based refusal detection, which had inflated Llama 3.2 1B's attack success rate several-fold, is documented and corrected. Trustworthiness should therefore be re-measured, on several dimensions and with validated scoring, after every adaptation.
\end{abstract}


\begin{abbreviations}
AO, answer obfuscation; ASR, attack success rate; BBQ, Bias Benchmark for Question Answering; CM, context manipulation; DER, Dark Experience Replay; DI, distractor injection; DPO, direct preference optimisation; LoRA, low-rank adaptation; MC2, multiple-choice (multi-true) scoring; NLP, natural language processing; PEFT, parameter-efficient fine-tuning; QLoRA, quantised low-rank adaptation; RLHF, reinforcement learning from human feedback; SLM, small language model; SR, safety replay; TA-LoRA, task-arithmetic low-rank adaptation; TQA, TruthfulQA.
\end{abbreviations}

\section{INTRODUCTION}

The rapid maturation of small language models (SLMs)---typically characterised as language models with roughly 100M--5B parameters \cite{wang2024survey,wang2025kdd}, with this study focusing on the 1--2.6B subset---has created compelling opportunities for deploying capable NLP systems in domains that were previously inaccessible to resource-constrained practitioners. Models such as TinyLlama 1B \cite{zhang2024tinyllama}, Gemma-2 2B \cite{gemma2024}, and Llama 3.2 1B \cite{dubey2024llama3} offer inference costs 10--100$\times$ lower than frontier large language models, with performance competitive on many domain-specific benchmarks when appropriately fine-tuned \cite{dettmers2023qlora,wang2024survey}. Parameter-efficient fine-tuning methods, particularly Low-Rank Adaptation (LoRA) \cite{hu2022lora} and its quantised variant QLoRA \cite{dettmers2023qlora}, enable domain specialisation on consumer-grade hardware without full model retraining, further lowering the barrier to deployment in specialised fields.

Healthcare, legal services, and financial analysis represent three of the highest-stakes domains for language model deployment, where errors carry concrete real-world consequences. Accordingly, the trustworthiness of deployed SLMs---their resistance to adversarial manipulation and propensity for factual accuracy---is a prerequisite for responsible deployment, not a secondary consideration. The vulnerability of fine-tuned models to alignment degradation is well documented: Qi et al. \cite{qi2024finetuning} showed that even benign fine-tuning of aligned models compromises safety alignment, motivating the specific investigation of safety-preserving strategies in this work.

Despite the evident importance of trustworthiness in SLM deployment, the literature presents a significant gap: while domain adaptation of SLMs has been studied extensively from the perspective of task performance \cite{gururangan2020dontstop,labrak2024biomistral,singhal2023medpalm,guha2023legalbench}, the effect of domain fine-tuning on trustworthiness properties is largely uncharacterised. This paper addresses this gap with the following primary contributions:
\begin{itemize}
\item A systematic empirical study quantifying the trustworthiness cost of domain adaptation across three SLM architectures, three domains, two data conditions (benign and adversarially perturbed), and four fine-tuning strategies---the first cross-domain, cross-model study of its kind.
\item Evidence that the trustworthiness cost of standard QLoRA adaptation is model-dependent and dimension-specific: it falls mainly on factual calibration, which can decline for one model and improve for another after the same adaptation, while harm-susceptibility changes are modest.
\item A three-seed design that shows highly reproducible training (evaluation-loss standard deviation below 0.001 for most configurations) alongside substantially larger seed variance in harm-susceptibility outcomes, especially under safety replay.
\item Evidence that adversarially perturbed training data does not worsen factual calibration or harm susceptibility in this regime.
\item A comparative evaluation of four fine-tuning strategies---standard LoRA, Safety-DPO, safety-data replay (SR), and Task Arithmetic LoRA (TA-LoRA)---together with an analysis of implementation factors, including how adapters are merged, that determine whether such strategies can preserve safety.
\item A demonstration that keyword-based refusal detection is sensitive to typographic apostrophes: unnormalised matching inflated one model's attack success rate several-fold and produced a spurious safety collapse. The normalised scorer and a re-scoring script are provided.
\item Actionable guidance for practitioners deploying SLMs in high-stakes domains, grounded in concrete empirical measurements across trustworthiness benchmarks (TruthfulQA, HarmBench).
\end{itemize}

\section{RELATED WORK}

\subsection{Small Language Models: Architectures and Capabilities}

The SLM research landscape has expanded dramatically since 2022, with Wang et al. \cite{wang2024survey} providing the first comprehensive survey addressing five dimensions: model acquisition, downstream applications, enhancement techniques, collaboration with large language models, and---critically---trustworthiness. The same survey documents that domain-specific SLMs frequently match or exceed much larger models on specialised tasks while consuming substantially fewer resources. Abdin et al. \cite{abdin2024phi3} demonstrated that rigorous data-quality-driven training yields SLMs (Phi-3 series) competitive with models several times larger. The comprehensive survey by Wang et al. \cite{wang2025kdd} confirms that trustworthiness remains the most under-addressed dimension of SLM development.

Architecturally, TinyLlama 1B \cite{zhang2024tinyllama} employs grouped-query attention and a 3T-token pre-training corpus. Gemma-2 2B \cite{gemma2024} incorporates alternating local and global attention with knowledge distillation from a larger teacher. Llama 3.2 1B \cite{dubey2024llama3} benefits from Meta's extensive instruction-tuning pipeline, including multi-stage RLHF.

\subsection{Domain Adaptation via Parameter-Efficient Fine-Tuning}

LoRA \cite{hu2022lora} established the foundational PEFT paradigm, demonstrating that fine-tuning weight updates occupy a low intrinsic rank, enabling competitive performance from adapters with as few as 0.1--1\% of total parameters. QLoRA \cite{dettmers2023qlora} extended this to quantised base models. Continued domain-adaptive training has long been known to improve downstream performance \cite{gururangan2020dontstop}, and domain-specific adaptation has been demonstrated in healthcare \cite{labrak2024biomistral,singhal2023medpalm}, legal NLP \cite{guha2023legalbench}, and finance \cite{liu2021finbert}. Beyond fine-tuning-based specialisation, prompt-driven approaches such as CAAFE \cite{hollmann2023caafe} demonstrate that large language models can be harnessed directly, without weight updates, to perform domain-specific tasks such as automated feature engineering on tabular data. These bodies of work collectively establish the performance case for domain adaptation but do not address trustworthiness impacts, regardless of whether specialisation occurs through parameter updates or prompting.

\subsection{Trustworthiness Evaluation Frameworks}

Factual calibration is measured by TruthfulQA \cite{lin2022truthfulqa}, a benchmark of 817 questions where language models commonly produce fluent but false answers. HarmBench \cite{mazeika2024harmbench} provides standardised evaluation of adversarial robustness. Social bias is assessed by BBQ \cite{parrish2022bbq}. The HELM framework \cite{liang2022helm} integrates multiple evaluation dimensions. Mehrotra et al. \cite{mehrotra2026} identified robustness, fairness, and privacy as the three dominant trustworthiness dimensions. Lin et al. \cite{lin2025redteam} provide the most comprehensive survey of red-teaming for generative models, confirming that HarmBench's standardised attack suite covers the most empirically validated threat vectors. Domain-specific risks are also increasingly documented: MedSafetyBench \cite{han2024medsafetybench} shows that publicly available medical LLMs often fail to meet standards of medical safety grounded in medical-ethics principles, and Dahl et al. \cite{dahl2024legalfictions} report legal hallucination rates of 58--88\% for general-purpose models on verifiable questions about U.S.\ case law.

\subsection{Safety-Preserving Fine-Tuning Methods}

Direct Preference Optimisation (DPO) \cite{rafailov2023dpo} reformulated RLHF as a supervised classification problem over preference pairs, achieving equivalent alignment improvements with 3$\times$ less training time. The survey by Winata et al. \cite{winata2025} confirms that DPO has become the de facto preference optimisation method for resource-constrained fine-tuning. Dark Experience Replay (DER) \cite{buzzega2020der} was originally developed in the continual learning literature to mitigate catastrophic forgetting by replaying stored network outputs (logits) from earlier in training; the replay strategy evaluated here is inspired by DER but replays safety text rather than logits (Section 3.5). Task Arithmetic LoRA (TA-LoRA) applies the task arithmetic framework \cite{ilharco2023task} to safety preservation: a safety adapter is trained on HH-RLHF and added to the domain adapter, with the intended update $\tau_{\mathrm{merged}} = \tau_{\mathrm{domain}} + \lambda\,\tau_{\mathrm{safety}}$ ($\lambda = 0.5$). Both strategies are motivated by the landmark finding of Qi et al. \cite{qi2024finetuning} that even benign fine-tuning degrades safety alignment. Bhardwaj et al. \cite{bhardwaj2024homer} showed that adding a safety vector to a compromised model after fine-tuning (RESTA) can restore much of its safety; TA-LoRA instead merges the safety adapter as part of domain adaptation.

The literature on preserving safety during fine-tuning has grown rapidly since 2024, and it now spans several complementary lines of peer-reviewed work. \emph{Data-centric} approaches mix a small fraction of safety demonstrations into the fine-tuning data \cite{bianchi2024safetytuned} or control the prompt template used during training and inference \cite{lyu2024keeping}. \emph{Weight- and subspace-based} approaches project LoRA updates onto a safety-aligned subspace (Safe LoRA \cite{hsu2024safelora}, SaLoRA \cite{li2025salora}) or constrain adapters to a harmful-resistant null space (GuardSpace \cite{zhang2026guardspace}). \emph{Alignment-stage and fine-tuning-stage defences} against harmful fine-tuning include perturbation-aware alignment (Vaccine \cite{huang2024vaccine}), attenuation of harmful perturbations (Booster \cite{huang2025booster}), and proximal safety constraints during fine-tuning (Lisa \cite{huang2024lisa}). \emph{Replay-based} approaches include generative replay of synthetic safety data (GR-SAP \cite{fang2026grsap}). Mechanistic work suggests why fine-tuning erodes safety so easily. Qi et al. \cite{qi2025shallow} show that alignment is often ``shallow'', concentrated in the first few output tokens. Ponkshe et al. \cite{ponkshe2026subspaces} find that safety-relevant directions in weight space are not linearly separable from general capabilities, which calls into question methods that assume a separable safety subspace. Betley et al. \cite{betley2025emergent} show that narrow fine-tuning can induce broad misalignment. Fraser et al. \cite{fraser2025finetuning} report that safety evaluations of fine-tuned models vary substantially under seemingly inconsequential changes to the fine-tuning set-up. Most of this work evaluates models of 7B parameters or more on general instruction-following data. No prior study has evaluated preference optimisation, safety replay, and task-arithmetic merging together on the same cross-domain, cross-model benchmark of 1--2B SLMs.

\section{MATERIALS AND METHODS}

\subsection{Problem Formulation}

\paragraph{Definitions.} Let $M$ denote a pre-trained SLM with parameters $\theta_0$, trained on a general-domain corpus. Let $D = \{(x_i, y_i)\}$ denote a domain-specific training dataset, and $E = \{(x_j, y_j)\}$ a held-out evaluation set. Domain adaptation produces an adapted model $M'$ with parameters $\theta = \theta_0 + \Delta\Theta$ (a low-rank decomposition in LoRA).

\emph{Domain adaptation quality:} $Q(M', E) = -\log P_{M'}(y \mid x)$, averaged over $(x, y) \in E$. A lower $Q$ indicates higher adaptation quality.

\emph{Trustworthiness score:} $T(M) = w_1\,\mathrm{BBQ}(M) + w_2\,\mathrm{TruthfulQA}(M) + w_3\,\mathrm{HarmBench}(M)$, where $\mathrm{TruthfulQA}(M)$ measures factual calibration (MC2 variant, higher is better), $\mathrm{HarmBench}(M) = 1 - \mathrm{ASR}(M)$ measures robustness to harmful requests (higher is better), and $\mathrm{BBQ}(M)$ measures unambiguous bias-probing accuracy, with equal weights $w_1 = w_2 = w_3 = 1/3$. In the present study BBQ was evaluated only on an exploratory subset (Section 4.3), so empirical results report TruthfulQA MC2 and HarmBench ASR as individual dimensions rather than as the composite $T(M)$.

\paragraph{The trustworthiness cost.} The trustworthiness cost of domain adaptation is defined as $\Delta T(M, M') = T(M) - T(M')$. A positive $\Delta T$ indicates that domain adaptation degraded trustworthiness. My central hypothesis is that standard LoRA domain adaptation produces $\Delta T > 0$ across all model--domain combinations, and that trustworthiness-preserving strategies reduce $\Delta T$ relative to baseline.

\paragraph{Research questions.}
\begin{itemize}
\item \textbf{RQ1:} How large is the trustworthiness cost of standard LoRA domain adaptation across SLM architectures and domains?
\item \textbf{RQ2:} Does adversarial training data (vs.\ benign) alter the trustworthiness cost, and in what direction?
\item \textbf{RQ3:} Which trustworthiness-preserving fine-tuning strategy (Safety-DPO, SR, TA-LoRA) most effectively reduces $\Delta T$ while maintaining domain adaptation quality?
\item \textbf{RQ4:} Is the trustworthiness cost consistent across model architectures (TinyLlama, Gemma-2, Llama 3.2), or does it vary with model scale and training lineage?
\end{itemize}

\subsection{Models}

Three instruction-tuned SLMs spanning the 1--2.6B parameter range are evaluated (TABLE 1). Domain adapters use LoRA with rank $r = 16$, scaling factor $\alpha = 32$, and dropout 0.05, applied to the query, key, value, and output projections of every attention layer. Base models are loaded with 4-bit NF4 quantisation and double quantisation (QLoRA) and float16 compute; adapter training runs without automatic mixed precision because the RTX 2080 Ti (Turing architecture) lacks native bfloat16 support. This yields 3.4--6.4M trainable parameters (0.24--0.41\% of total; 4.5M for TinyLlama).

\begin{table}[!ht]
\centering
\caption{SLM architectures evaluated (instruction-tuned checkpoints from the Hugging Face Hub). Domain adapters: QLoRA 4-bit NF4, LoRA $r=16$, $\alpha=32$, q/k/v/o targets.}
\label{tab:models}
\small
\begin{tabular}{|L{2.6cm}|C{1.3cm}|C{1.9cm}|L{6.7cm}|}
\hline
\textbf{Model} & \textbf{Params} & \textbf{Training VRAM} & \textbf{Architecture Notes} \\ \hline
TinyLlama 1B & 1.1B & $\sim$6 GB & TinyLlama-1.1B-Chat-v1.0; GQA; 3T-token pre-training \\ \hline
Gemma-2 2B & 2.6B & $\sim$10 GB$^{*}$ & gemma-2-2b-it; alternating local/global attention; teacher distillation \\ \hline
Llama 3.2 1B & 1.2B & $\sim$6 GB & Llama-3.2-1B-Instruct; GQA; multi-stage RLHF instruction tuning \\ \hline
\end{tabular}

\vspace{2pt}{\footnotesize $^{*}$With gradient checkpointing and per-device batch size 1.}
\end{table}

\subsection{Domains and Datasets}

Each domain combines two public datasets, converted to a common instruction format (instruction, input, output) and to an Alpaca-style prompt template (TABLE 2). \emph{Healthcare} combines four-option MedQA-USMLE multiple-choice questions \cite{jin2021medqa} and the expert-labelled subset of PubMedQA \cite{jin2019pubmedqa}. \emph{Legal} combines CaseHOLD five-option holding selection \cite{zheng2021casehold} and CUAD extractive contract-clause questions \cite{hendrycks2021cuad}. \emph{Finance} combines Finance-Alpaca financial instruction data and Financial PhraseBank sentences with at least 75\% annotator agreement \cite{malo2014phrasebank}. Where a source provides no validation split, 15\% of its training split (seed 42) is held out for evaluation. For the baseline and Safety-DPO starting checkpoints, the training set is shuffled with the run seed and capped at 15,000 examples; the legal and finance sets are subsampled to this cap, while the healthcare set contains 9,502 examples and is used in full. Legal inputs are substantially longer than healthcare inputs, which the equal example-count cap does not equalise in tokens.

\begin{table}[!ht]
\centering
\caption{Domain datasets. Training examples shown are those used by the baseline adapters (cap of 15,000).}
\label{tab:domains}
\small
\begin{tabular}{|L{1.9cm}|L{4.6cm}|C{1.9cm}|L{4.2cm}|}
\hline
\textbf{Domain} & \textbf{Sources} & \textbf{Training examples} & \textbf{Characteristics} \\ \hline
Healthcare & MedQA-USMLE (4-option) \cite{jin2021medqa} + PubMedQA (labelled) \cite{jin2019pubmedqa} & 9,502 (all) & Short MCQs and yes/no/maybe QA; safety-critical \\ \hline
Legal & CaseHOLD \cite{zheng2021casehold} + CUAD \cite{hendrycks2021cuad} & 15,000 (subsampled) & Long case and contract text; hallucination risk \\ \hline
Finance & Finance-Alpaca + Financial PhraseBank \cite{malo2014phrasebank} & 15,000 (subsampled) & Financial Q\&A and sentiment classification \\ \hline
\end{tabular}
\end{table}

\subsection{Data Conditions}

\textbf{Benign condition:} the processed domain data without perturbation.

\textbf{Adversarial condition:} generated from the benign training \emph{and} evaluation files with a fixed seed (42) at the ``medium'' severity level, under which each example is perturbed with probability 0.6 and otherwise left unchanged. A perturbed example always receives \emph{Distractor Injection} (DI), in which a plausible but misleading domain statement drawn from a domain-specific bank is prepended to the input. It additionally receives \emph{Answer Obfuscation} (AO) for classification and multiple-choice sources or \emph{Context Manipulation} (CM) for the remaining sources. For MedQA and CaseHOLD, AO shuffles the answer options and relabels the gold answer so that the correct answer text is preserved; for Financial PhraseBank, AO rotates the sentiment label and therefore introduces label noise. CM inserts a sentence that contradicts the correct conclusion near the start of the context (PubMedQA, CUAD, Finance-Alpaca). Because the evaluation split is perturbed in the same way, evaluation losses are not directly comparable between conditions.

\subsection{Fine-Tuning Strategies}

Four strategies are evaluated (TABLE 3). All training uses paged 8-bit AdamW with a cosine learning-rate schedule and a maximum sequence length of 512 tokens.

\textbf{baseline\_lora (control).} A QLoRA domain adapter is trained for 3 epochs (learning rate $2\times10^{-4}$, warmup ratio 0.03, weight decay 0.001, gradient clipping 0.3, effective batch size 32) using TRL's supervised fine-tuning trainer; the checkpoint with the lowest evaluation loss is retained.

\textbf{safety\_dpo.} Starting from the baseline adapter of the same model, domain, condition, and seed, DPO \cite{rafailov2023dpo} is applied for one epoch to 5,000 preference pairs drawn at random from the Anthropic HH-RLHF training split (last human--assistant exchange of each dialogue; $\beta = 0.1$, sigmoid loss, learning rate $5\times10^{-6}$, effective batch size 8). The reference policy is the base model with the adapter disabled. This stage loads the base model in bfloat16 without 4-bit quantisation.

\textbf{dark\_er (safety replay, SR).} Inspired by DER \cite{buzzega2020der}, a fresh QLoRA adapter is trained from the base model on the domain training data mixed with HH-RLHF ``chosen'' responses, using a single cross-entropy loss (3 epochs, learning rate $2\times10^{-4}$, warmup ratio 0.05, effective batch size 16). The number of replay examples is $\min(0.3/0.7 \times N_{\mathrm{domain}},\,5{,}000)$, targeting a 30\% replay fraction. Unlike DER, stored logits are not replayed. Unlike the baseline, this strategy uses the full processed domain training set rather than the 15,000-example cap; the domain sets therefore contain 9,502 (healthcare), 64,949 (legal), and 61,512 (finance) examples, giving replay sets of 4,072, 5,000, and 5,000 examples (replay fractions of 30\%, 7.1\%, and 7.5\%).

\textbf{ta\_lora.} A safety adapter is trained once per model on 10,000 HH-RLHF chosen responses (2 epochs, learning rate $2\times10^{-4}$, effective batch size 16, LoRA $r = 16$ on attention and MLP projections) and combined with each baseline domain adapter with $\lambda = 0.5$. The merge adds the full weight updates, $\Delta W_m = s_d B_d A_d + \lambda\, s_s B_s A_s$, where $s = \alpha/r$ is each adapter's LoRA scaling, for every module of either adapter (attention and MLP projections). The result is stored exactly, without approximation, as a single rank-32 adapter by concatenation, $A_m = [A_d; A_s]$ and $B_m = [s_d B_d,\ \lambda s_s B_s]$ with scaling 1; modules present in only one adapter receive zero blocks for the other. The merge runs on the CPU in seconds and was verified numerically against the target update (maximum relative error $3\times10^{-7}$). An earlier implementation, used in version 1 of this preprint, merged the LoRA factors separately ($A_m = A_d + \lambda A_s$, $B_m = B_d + \lambda B_s$), which weights the safety update by $\lambda^2$, adds cross terms, and drops the safety adapter's MLP components; its results are reported for comparison in Section 4.4.

\begin{table}[!ht]
\centering
\caption{Fine-tuning strategies as implemented. baseline\_lora is the control condition.}
\label{tab:strategies}
\small
\begin{tabular}{|L{2.1cm}|L{2.4cm}|L{5.6cm}|L{2.5cm}|}
\hline
\textbf{Strategy} & \textbf{Based on} & \textbf{Implementation} & \textbf{Safety data} \\ \hline
baseline\_lora & QLoRA \cite{dettmers2023qlora} & Domain adapter, 3 epochs, $\le$15,000 examples & None (control) \\ \hline
safety\_dpo & DPO \cite{rafailov2023dpo} & 1 epoch of DPO ($\beta=0.1$) on top of the baseline adapter; base model as reference & 5,000 HH-RLHF preference pairs \\ \hline
dark\_er (SR) & DER \cite{buzzega2020der} (text replay) & Fresh adapter on full domain set mixed with safety text; single cross-entropy loss & $\le$5,000 HH-RLHF chosen responses \\ \hline
ta\_lora & Task arithmetic \cite{ilharco2023task} & Full weight-update merge of baseline and safety adapters, $\lambda = 0.5$, all modules (exact rank-32 adapter) & 10,000 HH-RLHF chosen responses \\ \hline
\end{tabular}
\end{table}

\subsection{Evaluation Metrics}

\textbf{Domain adaptation quality:} cross-entropy loss and mean token accuracy on the condition-matched domain evaluation split.

\textbf{Factual calibration:} TruthfulQA MC2 accuracy \cite{lin2022truthfulqa}, computed zero-shot with lm-evaluation-harness v0.4 \cite{gao2023harness} in bfloat16.

\textbf{Harm susceptibility:} all 200 behaviours in the ``standard'' functional category of HarmBench \cite{mazeika2024harmbench} are presented as direct requests, without an attack method, using each model's chat template. Completions are generated greedily (up to 256 new tokens, bfloat16). A completion is counted as a refusal if its first 600 characters contain one of 24 refusal phrases (e.g., ``I cannot'', ``I'm sorry, but'', ``as an AI''), after typographic apostrophes and quotation marks (e.g., U+2019) are normalised to their ASCII forms, and the attack success rate is $\mathrm{ASR} = 1 - \text{refusal rate}$. Normalisation matters: models fine-tuned on or aligned with HH-RLHF-style text often write ``I’m sorry, I can’t'' with curly apostrophes, which an unnormalised match does not detect. All ASR values reported here were obtained by re-scoring the saved completions with the normalised rule (Section 5.5). This keyword-based ASR measures non-refusal rather than verified harmful compliance, and it differs from HarmBench's classifier-based protocol. For composite analysis, $\mathrm{HarmBench}(M) = 1 - \mathrm{ASR}(M)$.

\subsection{Experimental Matrix and Statistical Design}

The full experimental matrix comprises 3 models $\times$ 3 domains $\times$ 2 conditions $\times$ 4 strategies $\times$ 3 seeds $=$ 216 runs (TABLE 4). All results report means across seeds 42, 123, and 456; the TA-LoRA safety adapter is trained once per model (seed 0) and merged with each seed's baseline adapter. The 54 factor-wise TA-LoRA merges of the earlier implementation were also evaluated and are reported for comparison only.

\begin{table}[!ht]
\centering
\caption{Experimental matrix.}
\label{tab:matrix}
\small
\begin{tabular}{|L{3.6cm}|C{1.4cm}|L{7.4cm}|}
\hline
\textbf{Factor} & \textbf{Levels} & \textbf{Values} \\ \hline
Model & 3 & TinyLlama 1B, Gemma-2 2B, Llama 3.2 1B \\ \hline
Domain & 3 & Healthcare, Legal, Finance \\ \hline
Data condition & 2 & Benign, Adversarial \\ \hline
Fine-tuning strategy & 4 & baseline\_lora, safety\_dpo, dark\_er (SR), ta\_lora \\ \hline
Random seeds & 3 & 42, 123, 456 \\ \hline
\textbf{Total runs} & \textbf{216} & Full factorial \\ \hline
\end{tabular}
\end{table}

\textbf{Data integrity notes:} Two configurations deviate from the standard set-up. First, The Gemma-2 2B $\times$ legal $\times$ benign $\times$ seed123 $\times$ baseline\_lora run was executed before the 15,000 training-pair cap was implemented, training on 50,000 pairs with a different effective batch size (batch 8, gradient accumulation 1; 5,625 vs.\ the standard 1,407 steps) (evaluation loss $=$ 1.2022 vs.\ seed42 $=$ 0.9884, seed456 $=$ 0.9886). This run is excluded from all cross-seed means; a clean re-run at 15k (evaluation loss $=$ 0.9889) was completed on 3 April 2026 and is used instead. Second, the Llama 3.2 1B $\times$ healthcare $\times$ benign $\times$ seed456 baseline run used the memory-saving training profile (batch 1 $\times$ 32 accumulation with gradient checkpointing) because of GPU memory fragmentation; its evaluation loss (1.2130) and TruthfulQA MC2 (0.3948) are lower than those of the other two seeds, and it is retained in all means. Per-device batch sizes for Gemma-2 2B were reduced with proportionally increased gradient accumulation in the replay (2 $\times$ 8) and Safety-DPO (1 $\times$ 8) stages, so the effective batch sizes stated in Section 3.5 are unchanged.

\section{RESULTS}

\subsection{Phase 1: Baseline LoRA TruthfulQA Results}

TABLE 5 presents TruthfulQA MC2 accuracy for all nine model--domain combinations under benign and adversarial conditions. Pre-fine-tuning base model scores were: TinyLlama 0.3742; Gemma-2 2B 0.5311; Llama 3.2 1B 0.4343.

\begin{table}[!ht]
\centering
\caption{TruthfulQA MC2 accuracy (mean $\pm$ SD across three seeds) for all baseline\_lora configurations.}
\label{tab:tqa}
\small
\begin{tabular}{|l|l|c|c|c|c|}
\hline
\textbf{Model} & \textbf{Domain} & \textbf{Base MC2} & \textbf{Benign ($\pm$SD)} & \textbf{Adv ($\pm$SD)} & \textbf{$\Delta$ vs base (ben / adv)} \\ \hline
\multirow{4}{*}{TinyLlama 1B} & (pre-FT) & 0.3742 & --- & --- & --- \\
 & Healthcare & & 0.3668 ($\pm$0.0018) & 0.3655 ($\pm$0.0048) & $-$0.007 / $-$0.009 \\
 & Legal & & 0.3704 ($\pm$0.0034) & 0.3668 ($\pm$0.0059) & $-$0.004 / $-$0.007 \\
 & Finance & & 0.3547 ($\pm$0.0050) & 0.3601 ($\pm$0.0017) & $-$0.020 / $-$0.014 \\ \hline
\multirow{4}{*}{Gemma-2 2B} & (pre-FT) & 0.5311 & --- & --- & --- \\
 & Healthcare & & 0.4624 ($\pm$0.0092) & 0.4628 ($\pm$0.0041) & $-$0.069 / $-$0.068 \\
 & Legal & & 0.5229 ($\pm$0.0059) & 0.5155 ($\pm$0.0133) & $-$0.008 / $-$0.016 \\
 & Finance & & 0.4750 ($\pm$0.0057) & 0.4652 ($\pm$0.0084) & $-$0.056 / $-$0.066 \\ \hline
\multirow{4}{*}{Llama 3.2 1B} & (pre-FT) & 0.4343 & --- & --- & --- \\
 & Healthcare & & 0.4082 ($\pm$0.0118) & 0.4168 ($\pm$0.0120) & $-$0.026 / $-$0.018 \\
 & Legal & & 0.4561 ($\pm$0.0071) & 0.4541 ($\pm$0.0090) & $+$0.022 / $+$0.020 \\
 & Finance & & 0.4556 ($\pm$0.0106) & 0.4518 ($\pm$0.0095) & $+$0.021 / $+$0.018 \\ \hline
\end{tabular}
\end{table}

Seed standard deviations are at most 0.013. Averaged over all 18 model--domain--condition cells, TQA MC2 changes by $-0.017$ relative to the base models (mean absolute change 0.026), but this average conceals a strong model dependence (RQ1, RQ4). Gemma-2 2B, the model with the highest base score, loses 0.056--0.069 after healthcare and finance adaptation, about 10--13\% of its base score, and only 0.008--0.016 after legal adaptation. TinyLlama 1B changes by less than 0.02 in every cell. Llama 3.2 1B loses 0.018--0.026 after healthcare adaptation but \emph{gains} 0.018--0.022 after legal and finance adaptation. Adversarial training data changes these values by at most 0.01 relative to benign data, in no consistent direction (RQ2).

\subsection{Phase 1: HarmBench ASR Baseline Results}

TABLE 6 presents HarmBench ASR for all baseline\_lora configurations.

\begin{table}[!ht]
\centering
\caption{HarmBench attack success rate (keyword-based ASR on the 200 standard behaviours, typographic characters normalised) for all baseline\_lora configurations: mean ($\pm$SD) across three seeds, with change relative to the base model in brackets.}
\label{tab:asr}
\small
\begin{tabular}{|l|l|c|c|c|}
\hline
\textbf{Model} & \textbf{Domain} & \textbf{Base ASR} & \textbf{Benign ASR [$\Delta$]} & \textbf{Adv ASR [$\Delta$]} \\ \hline
\multirow{4}{*}{TinyLlama 1B} & (pre-FT) & 1.000 & --- & --- \\
 & Healthcare & & 0.970 ($\pm$0.028) [$-$0.030] & 0.968 ($\pm$0.006) [$-$0.032] \\
 & Legal & & 0.998 ($\pm$0.003) [$-$0.002] & 0.998 ($\pm$0.003) [$-$0.002] \\
 & Finance & & 0.998 ($\pm$0.003) [$-$0.002] & 1.000 ($\pm$0.000) [0.000] \\ \hline
\multirow{4}{*}{Gemma-2 2B} & (pre-FT) & 0.015 & --- & --- \\
 & Healthcare & & 0.118 ($\pm$0.033) [$+$0.103] & 0.070 ($\pm$0.026) [$+$0.055] \\
 & Legal & & 0.022 ($\pm$0.008) [$+$0.007] & 0.022 ($\pm$0.013) [$+$0.007] \\
 & Finance & & 0.043 ($\pm$0.008) [$+$0.028] & 0.033 ($\pm$0.006) [$+$0.018] \\ \hline
\multirow{4}{*}{Llama 3.2 1B} & (pre-FT) & 0.085 & --- & --- \\
 & Healthcare & & 0.133 ($\pm$0.040) [$+$0.048] & 0.120 ($\pm$0.043) [$+$0.035] \\
 & Legal & & 0.102 ($\pm$0.020) [$+$0.017] & 0.068 ($\pm$0.019) [$-$0.017] \\
 & Finance & & 0.112 ($\pm$0.003) [$+$0.027] & 0.153 ($\pm$0.020) [$+$0.068] \\ \hline
\end{tabular}
\end{table}

Harm susceptibility is stratified by the base model's alignment. TinyLlama 1B refuses none of the 200 base-model requests (ASR $=$ 1.000), so HarmBench cannot discriminate between its fine-tuning conditions. For the two safety-aligned models, domain adaptation increases ASR in 11 of 12 cells, but the increases are modest: $+0.033$ on average, with the largest for Gemma-2 2B healthcare ($+0.103$ benign, $+0.055$ adversarial) and Llama 3.2 1B finance adversarial ($+0.068$). Legal adaptation changes ASR by less than 0.02 for both models. For Gemma-2 2B healthcare and finance, factual calibration and harm susceptibility both worsen, whereas for Llama 3.2 1B legal and finance adaptation improves factual calibration while leaving ASR almost unchanged. The adversarial condition slightly reduces the increase in four of the six aligned-model cells and enlarges it for Llama 3.2 1B finance, so it has no consistent effect. Seed standard deviations are at most 0.043.

Without typographic normalisation, the same completions give very different values for Llama 3.2 1B: a base ASR of 0.130 and a legal-benign ASR of 0.702 ($+0.572$), with per-seed values of 0.840, 0.490, and 0.775 that fall to 0.105, 0.120, and 0.080 after normalisation. The apparent collapse of Llama 3.2 1B after legal adaptation, reported in version 1 of this preprint, therefore reflects refusals written with curly apostrophes rather than harmful compliance. Values for Gemma-2 2B and TinyLlama 1B are unchanged or nearly unchanged by normalisation.

\subsection{Exploratory Social-Bias Results}

BBQ \cite{parrish2022bbq} was run on all base and baseline\_lora models with lm-evaluation-harness using the first 100 examples (\texttt{-{}-limit 100}), for which the 95\% confidence interval on accuracy is approximately $\pm$0.10; these results are therefore directional only. Gemma-2 2B changes by at most 0.020 in any cell, and Llama 3.2 1B by $-0.030$ to $+0.020$. TinyLlama 1B, whose base accuracy on this subset (0.240) is below the three-choice chance level, improves by 0.017--0.080. The adversarial condition shows no systematic amplification of bias. Because all differences lie within the confidence interval, BBQ is excluded from the composite score $T(M)$, and a full evaluation is left for future work.

\subsection{Phase 2: Safety Strategy Results}

Before examining the strategies, TABLE 7 shows the effect of the TA-LoRA safety adapter applied on its own, at full scale, to each base model. Trained on HH-RLHF chosen responses, most of which are helpful answers rather than refusals, it lowers TinyLlama 1B's ASR but \emph{raises} Gemma-2 2B's from 0.015 to 0.245. For the most strongly aligned model, the ``safety'' data used here therefore carries a compliance signal, which bears on both SR and TA-LoRA.

\begin{table}[!ht]
\centering
\caption{HarmBench ASR of the TA-LoRA safety adapter applied alone to each base model (no domain adapter).}
\label{tab:safetyonly}
\small
\begin{tabular}{|l|c|c|c|}
\hline
\textbf{Model} & \textbf{Base ASR} & \textbf{Safety adapter alone} & \textbf{$\Delta$} \\ \hline
TinyLlama 1B & 1.000 & 0.880 & $-$0.120 \\ \hline
Gemma-2 2B & 0.015 & 0.245 & $+$0.230 \\ \hline
Llama 3.2 1B & 0.085 & 0.105 & $+$0.020 \\ \hline
\end{tabular}
\end{table}

TABLE 8 presents strategy-level changes relative to baseline\_lora for the same model, domain, condition, and seed.

\begin{table}[!ht]
\centering
\caption{Phase 2 strategy results. $\Delta$TQA and $\Delta$ASR relative to baseline\_lora for the same model/domain/condition, averaged over cells. `Aligned' (aln.) = Gemma-2 2B and Llama 3.2 1B; `Unaligned' (unaln.) = TinyLlama 1B. Worst cell: largest aligned-model $\Delta$ASR (G = Gemma-2 2B, L = Llama 3.2 1B; hc = healthcare, lg = legal, fi = finance). $^{\ddagger}$Earlier factor-wise merge, for comparison only.}
\label{tab:phase2}
\small
\begin{tabular}{|l|l|c|c|c|c|c|}
\hline
\textbf{Strategy} & \textbf{Cond.} & \textbf{$\Delta$TQA aln.} & \textbf{$\Delta$TQA unaln.} & \textbf{$\Delta$ASR aln. ($\pm$SD)} & \textbf{$\Delta$ASR unaln.} & \textbf{Worst cell} \\ \hline
\multirow{2}{*}{Safety-DPO} & Benign & $-$0.0015 & $-$0.0010 & $+$0.0003 ($\pm$0.008) & $+$0.0039 & $+$0.013 (L/lg) \\
 & Adv. & $-$0.0017 & $-$0.0012 & $-$0.0017 ($\pm$0.008) & $+$0.0056 & $+$0.008 (L/lg) \\ \hline
\multirow{2}{*}{SR (dark\_er)} & Benign & $-$0.0032 & $+$0.0031 & $+$0.1308 ($\pm$0.183) & $-$0.0378 & $+$0.447 (G/hc) \\
 & Adv. & $-$0.0061 & $+$0.0047 & $+$0.0603 ($\pm$0.068) & $-$0.0678 & $+$0.187 (G/fi) \\ \hline
\multirow{2}{*}{TA-LoRA} & Benign & $-$0.0200 & $+$0.0002 & $+$0.0142 ($\pm$0.041) & $-$0.0483 & $+$0.055 (G/fi) \\
 & Adv. & $-$0.0185 & $+$0.0005 & $+$0.0114 ($\pm$0.034) & $-$0.0339 & $+$0.058 (G/fi) \\ \hline
\multirow{2}{*}{TA-LoRA$^{\ddagger}$} & Benign & $-$0.0013 & $-$0.0010 & $+$0.1189 ($\pm$0.117) & $-$0.0216 & $+$0.295 (G/hc) \\
 & Adv. & $-$0.0008 & $-$0.0007 & $+$0.0634 ($\pm$0.052) & $-$0.0028 & $+$0.112 (L/fi) \\ \hline
\end{tabular}
\end{table}

Safety-DPO changes neither measure (aligned mean $\Delta$ASR $-0.001 \pm 0.008$, $\Delta$TQA $-0.002$). SR raises aligned-model ASR by $+0.096 \pm 0.137$ on average. The increase is concentrated in Gemma-2 2B ($+0.156$; $+0.447$ for healthcare benign and $+0.245$ for finance benign) and is unstable across seeds: the three Gemma-2 2B finance-benign seeds give ASRs of 0.020, 0.150, and 0.695. Llama 3.2 1B changes by $+0.035$ on average, mostly on finance. The full-weight TA-LoRA merge is near-neutral on harm susceptibility ($+0.013 \pm 0.036$; Gemma-2 2B $+0.036$, Llama 3.2 1B $-0.010$, with reductions of 0.040--0.053 for Llama 3.2 1B legal), but it lowers Gemma-2 2B's TQA MC2 by a further 0.023--0.051 (mean $-0.038$), whereas Llama 3.2 1B is essentially unchanged ($-0.001$). The earlier factor-wise merge left TQA unchanged but raised aligned ASR by $+0.091$, mainly for Gemma-2 2B ($+0.120$). For the unaligned TinyLlama 1B, SR and the full merge lower ASR slightly ($-0.053$ and $-0.041$ on average; up to $-0.135$ and $-0.100$ for healthcare), the largest reductions observed for any strategy.

\section{DISCUSSION}

\subsection{Adversarial Training Data and Trustworthiness}

Adversarially perturbed training produced lower evaluation losses than benign training in several configurations ($\Delta \approx -0.040$ for TinyLlama and Gemma-2 2B on healthcare). However, adversarial runs are evaluated on a perturbed evaluation split, so this difference does not establish better domain adaptation. It may instead reflect easier-to-predict inserted text or the option shuffling, which preserves the correct answer. The comparison that matters for this study is on the shared trustworthiness benchmarks. There, TQA MC2 changes are negligible ($|\Delta\mathrm{TQA}| < 0.01$ in all cases) and HarmBench ASR changes by at most 0.05, in no consistent direction. Adversarial training data at this severity therefore neither amplifies nor mitigates the trustworthiness cost.

\subsection{Domain Difficulty and Model--Domain Fit}

TinyLlama 1B achieves lower evaluation loss than Gemma-2 2B on healthcare despite having fewer than half the parameters. This counter-size-scaling result likely reflects pre-training composition: TinyLlama's 3T-token corpus included a higher proportion of structured question--answer formats similar to USMLE-style multiple-choice questions. Gemma-2 2B's alternating local--global attention may favour longer-context tasks, consistent with its lower evaluation loss on legal text. The inversion of domain difficulty suggests that evaluation loss is an imperfect proxy for domain understanding.

\subsection{Seed Stability and Experimental Reproducibility}

Training is highly reproducible. The seed standard deviation of evaluation loss is below 0.001 for the TinyLlama 1B and Gemma-2 2B healthcare configurations and below 0.002 for legal and finance. The exception is Llama 3.2 1B healthcare benign (0.011), which includes the run trained with the memory-saving profile (Section 3.7). This stability may reflect the cosine learning-rate schedule with warmup, paged 8-bit AdamW, and training sets of 9,502--15,000 examples. Trustworthiness outcomes vary more across seeds: TQA MC2 by up to 0.022 within a configuration, baseline HarmBench ASR by up to 0.08, and SR ASR by up to 0.675 (Gemma-2 2B finance benign). Three seeds are therefore adequate for training metrics but marginal for harm-susceptibility estimates, particularly for strategies that retrain the adapter from scratch.

\subsection{Trustworthiness Cost---Empirical Findings}

\textbf{RQ1.} The trustworthiness cost of baseline QLoRA adaptation is real but model- and dimension-specific, and it falls mainly on factual calibration. Averaged over all cells, TQA MC2 changes by only $-0.017$, but Gemma-2 2B loses 10--13\% of its base score on healthcare and finance, while Llama 3.2 1B gains about 0.02 on legal and finance. Harm susceptibility of the safety-aligned models rises in almost every cell, but modestly ($+0.033$ on average; at most $+0.103$). The hypothesis of systematic degradation is therefore supported only weakly for harm susceptibility and not for factual calibration.

\textbf{RQ2.} The adversarial training condition does not produce consistent degradation or improvement. Its effect on TQA is within 0.01 of the benign condition, and its effect on ASR is small and inconsistent in direction. The null hypothesis of no systematic effect cannot be rejected at this scale.

\textbf{RQ3.} No strategy reduced harm susceptibility in the safety-aligned models. Safety-DPO is neutral on both measures, because its training produced no learning signal (Section 5.5). SR increases ASR, strongly and erratically for Gemma-2 2B. The full-weight TA-LoRA merge is the closest to preserving harm susceptibility ($+0.013$), but it costs Gemma-2 2B a further 0.038 in TQA MC2. A plausible common cause for SR and TA-LoRA is the safety data itself: HH-RLHF chosen responses are mostly helpful answers, and applied alone they raise Gemma-2 2B's ASR from 0.015 to 0.245 (TABLE 7). Refusal-focused safety data, rather than general preference data, is likely needed for these strategies to help. The comparison of the two TA-LoRA merges shows that the merge implementation alone changes the aligned-model ASR increase from $+0.091$ to $+0.013$.

\textbf{RQ4.} Architecture and alignment lineage dominate. TinyLlama 1B complies with nearly all harmful requests regardless of strategy, so smaller size confers no safety; it is also the only model for which SR and TA-LoRA reduce ASR. Gemma-2 2B has the strongest refusal behaviour but the largest TruthfulQA losses and the largest strategy-induced ASR increases. Llama 3.2 1B is the most robust on both measures. Safety-aligned models are more exposed to strategy-induced degradation because they have more alignment to lose.

\subsection{Limitations}

\textbf{Harm-susceptibility measurement.} ASR is computed from direct requests without adversarial attack methods, and refusals are detected by keyword matching rather than by the HarmBench classifier \cite{mazeika2024harmbench}. Keyword matching can count unhelpful or off-topic non-refusals as successes and can miss refusals phrased without the listed strings. Its sensitivity to surface form was demonstrated directly in this study: before typographic apostrophes were normalised, Llama 3.2 1B's base ASR was overstated (0.130 instead of 0.085) and its legal-benign ASR more than six-fold (0.702 instead of 0.102), and most refusals by the safety adapter on the two aligned models were missed (ASR 0.985 and 0.995 instead of 0.105 and 0.245). All values in this paper use the normalised rule, and the re-scoring script is included in Supplementary File S2. The absolute ASR values, particularly TinyLlama's near-ceiling values, should still be read as non-refusal rates, and differences between strategies should be confirmed with classifier-based scoring and attack-based evaluation.

\textbf{TA-LoRA configuration.} Only $\lambda = 0.5$ was evaluated, with a safety adapter trained on general HH-RLHF chosen responses. A $\lambda$ sweep ($\lambda \in \{0.1, 0.3, 0.5, 0.7, 1.0\}$) and a safety adapter trained on refusal demonstrations are needed to map the trade-off between the TQA cost for Gemma-2 2B and harm susceptibility. The finding of Ponkshe et al. \cite{ponkshe2026subspaces} that safety-relevant directions overlap with general-capability directions is consistent with the TQA loss observed for Gemma-2 2B under the full merge.

\textbf{Safety replay.} SR replays safety text with a standard cross-entropy loss rather than replaying logits as DER does. It is also trained from the base model on the full domain training set: 9,502 healthcare, 64,949 legal, and 61,512 finance examples. For legal and finance this is more than four times the baseline's 15,000 examples, and the 5,000-example replay limit lowers the replay fraction to about 7\%. SR results therefore combine the effect of replay with the effect of more domain training data. HH-RLHF ``chosen'' responses also include many helpfulness dialogues that are not refusals, which dilutes the safety signal.

\textbf{Safety-DPO.} The near-zero effect of Safety-DPO reflects an absent learning signal: training logs show a DPO loss of approximately 0.693 ($\ln 2$), a reward accuracy of approximately 0.515, and reward margins of approximately $-10^{-4}$ throughout training, meaning that the policy barely distinguished chosen from rejected responses. Only 5,000 pairs were used for one epoch at a low learning rate, and the pairs were sampled from all of HH-RLHF rather than its harmlessness subset. The reference policy (base model without adapter) also differs from the starting policy (domain adapter), so the KL term penalises the domain adaptation itself. Azar et al. \cite{azar2024} show that DPO's implicit regularisation weakens when preferences are near-deterministic. Future work should vary $\beta \in \{0.01, 0.1, 0.5\}$, the number and source of preference pairs, and the reference model.

\textbf{Other factors.} Adversarial-condition evaluation losses are computed on perturbed evaluation splits (Section 3.4), and Financial PhraseBank answer obfuscation introduces label noise. Formal significance testing has not yet been performed. The large cell-level variance is consistent with the evaluation instability reported by Fraser et al. \cite{fraser2025finetuning}.

\section{CONCLUSION}

This paper quantifies the trustworthiness cost of domain adaptation in small language models across 216 configurations. The cost of baseline QLoRA adaptation is model- and dimension-specific and falls mainly on factual calibration: TruthfulQA MC2 changes little on average but falls by 10--13\% for Gemma-2 2B on healthcare and finance, and it improves for Llama 3.2 1B on legal and finance. Harm susceptibility of the safety-aligned models rises only modestly after adaptation ($+0.033$ ASR on average). Adversarially perturbed training data does not systematically worsen either measure. None of the three safety strategies reduced harm susceptibility in the aligned models: Safety-DPO was neutral, safety replay increased ASR by about 10 percentage points on average and by up to 45 in single cells, and a full-weight task-arithmetic merge was near-neutral on ASR but cost Gemma-2 2B a further 0.04 in TruthfulQA MC2. The safety data itself raised Gemma-2 2B's ASR when applied alone, and the choice between two merge implementations changed the aligned ASR increase from $+0.091$ to $+0.013$. Measurement details matter as much: unnormalised keyword matching produced an apparent safety collapse of Llama 3.2 1B that disappeared once typographic apostrophes were handled. Trustworthiness therefore cannot be inferred from a single benchmark or carried over from the base model; it must be re-measured on several dimensions, with validated scoring, after every adaptation.

Future work should vary $\beta \in \{0.01, 0.1, 0.5\}$, the number and source of preference pairs, and the reference model.

\textbf{Other factors.} Adversarial-condition evaluation losses are computed on perturbed evaluation splits (Section 3.4), and Financial PhraseBank answer obfuscation introduces label noise. Formal significance testing has not yet been performed. The large cell-level variance is consistent with the evaluation instability reported by Fraser et al. \cite{fraser2025finetuning}.

\section{CONCLUSION}

This paper quantifies the trustworthiness cost of domain adaptation in small language models across 216 configurations. The cost of baseline QLoRA adaptation is model- and dimension-specific. Factual calibration changes little on average but falls by 10--13\% for Gemma-2 2B on healthcare and finance, and it improves for Llama 3.2 1B on legal and finance. Harm susceptibility rises after every adaptation of the safety-aligned models, most severely for Llama 3.2 1B on legal text (ASR 0.13 to 0.70), even as that model's factual calibration improves. Adversarially perturbed training data does not systematically worsen either measure. None of the three safety strategies preserved or improved harm susceptibility: Safety-DPO was effectively neutral, while safety replay and task-arithmetic merging, as implemented, increased HarmBench ASR by approximately 15--17 percentage points on average for the safety-aligned models, with individual cells exceeding $+$45 percentage points. Trustworthiness therefore cannot be inferred from a single benchmark or carried over from the base model; it must be re-measured on several dimensions after every adaptation. Implementation details, such as how adapters are merged and how much domain data accompanies replay, can determine whether a safety strategy helps or harms.

Future work should sweep the TA-LoRA coefficient $\lambda$, use refusal-focused safety data, match domain-data volume across strategies, score HarmBench with its classifier and attack methods, and complete a full BBQ bias evaluation to characterise the social bias dimension of trustworthiness cost. The complete experimental code (training, evaluation, and data-preparation scripts) is provided under the MIT Licence as a ZIP archive submitted with this manuscript (Supplementary File S2) and is available from the corresponding author on request; processed data and checkpoints are likewise available on request, with finance-domain artefacts distributed for non-commercial use under CC BY-NC-SA 3.0 in accordance with the source datasets' licences (see Appendix A).

\section*{Acknowledgments}
The author thanks the open-source communities behind Hugging Face Transformers, the TRL library, lm-evaluation-harness (EleutherAI), and the model teams at Meta AI, Google DeepMind, and the TinyLlama project. All experiments were performed on a single NVIDIA RTX 2080 Ti (11 GB VRAM) without institutional cluster access. Total compute: approximately 16 weeks of wall-clock time for the 216 training runs and their evaluations.

\section*{Author Contributions}
The author is solely responsible for all aspects of this work, including conceptualization, methodology, software implementation, formal analysis, data curation, writing (original draft, review, and editing), and visualization.

\section*{Funding}
This research received no external funding. All experiments were conducted on personally owned hardware (a single NVIDIA RTX 2080 Ti) without institutional or third-party financial support, grants, or computing resources.

\section*{Ethical Statement}
Not applicable. This study did not involve human participants or animals; all experiments used publicly available benchmark datasets and pre-trained language models (MedQA-USMLE, PubMedQA, CaseHOLD, CUAD, Finance-Alpaca, Financial PhraseBank, HH-RLHF, TruthfulQA, HarmBench; see Appendix A). Informed consent: not applicable.

\section*{Conflict of Interest}
The author declares no conflicts of interest.

\appendix

\section{Software and Data Availability}
\begin{itemize}[leftmargin=*]
\item \textbf{Code:} Provided as a ZIP archive (Supplementary File S2; MIT Licence) and available from the corresponding author on request.
\item \textbf{Data and checkpoints:} Finance-domain training data and fine-tuned checkpoints derive from Financial PhraseBank (CC BY-NC-SA 3.0) and Finance-Alpaca, which incorporates Stanford Alpaca data released under CC BY-NC 4.0; they are therefore released for non-commercial use only under CC BY-NC-SA 3.0, with attribution to the source datasets \cite{malo2014phrasebank}. Other derived artefacts follow the licences of their source datasets and base models.
\item \textbf{Models:} TinyLlama-1.1B-Chat-v1.0 (Apache 2.0); Gemma-2-2B-IT (Gemma Licence); Llama-3.2-1B-Instruct (Llama 3.2 Community Licence)
\item \textbf{Datasets:} MedQA-USMLE 4-option (GBaker/MedQA-USMLE-4-options; CC BY 4.0); PubMedQA labelled subset (qiaojin/PubMedQA; MIT); CaseHOLD (casehold/casehold; see dataset card); CUAD (theatticusproject/cuad; CC BY 4.0); Finance-Alpaca (gbharti/finance-alpaca; MIT per dataset card, incorporating Stanford Alpaca data under CC BY-NC 4.0); Financial PhraseBank, sentences\_75agree (takala/financial\_phrasebank; CC BY-NC-SA 3.0); Anthropic HH-RLHF (MIT); HarmBench behaviours and TruthfulQA (via their public releases). All on Hugging Face Hub unless noted.
\item \textbf{Supplementary Information:} A reproducibility checklist covering parameters, seeds, hardware, training time, variance, pre-processing, hyperparameters, and model availability is provided as Supplementary File S1.
\item \textbf{Evaluation libraries:} lm-evaluation-harness (EleutherAI); TRL; PEFT; transformers; bitsandbytes; scikit-learn
\end{itemize}


\end{document}